\documentclass{bmvc2k}
\usepackage{framed,multirow}
\usepackage{cleveref}
\title{Clustered Saliency Prediction}

\addauthor{Rezvan Sherkati}{rezvan.sherkati@mail.mcgill.ca}{1}
\addauthor{James J. Clark}{james.clark1@mcgill.ca}{1}

\addinstitution{
 Electrical and Computer Engineering Department,\\ McGill University\\ 845 Sherbrooke St W,\\ Montreal, Canada
}

\runninghead{Rezvan Sherkati, James J. Clark}{Clustered Saliency Prediction}

\def\etal{\emph{et al}\bmvaOneDot}

\begin{document}

\maketitle

\begin{abstract}
We present a new method for image salience prediction, Clustered Saliency Prediction. This method divides subjects into clusters based on their personal features and their known saliency maps, and generates an image salience model conditioned on the cluster label. We test our approach on a public dataset of personalized saliency maps and cluster the subjects using selected importance weights for personal feature factors. We propose the Multi-Domain Saliency Translation model which uses image stimuli and universal saliency maps to predict saliency maps for each cluster. For obtaining universal saliency maps, we applied various state-of-the-art methods, DeepGaze IIE, ML-Net and SalGAN, and compared their effectiveness in our system. We show that our Clustered Saliency Prediction technique outperforms the universal saliency prediction models. Also, we demonstrate the effectiveness of our clustering method by comparing the results of Clustered Saliency Prediction using clusters obtained by our algorithm with some baseline methods. Finally, we propose an approach to assign new people to their most appropriate cluster and prove its usefulness in the experiments.
\end{abstract}

\section{Introduction}
When humans look at an image, they might fixate on some points (i.e. fixation points). Fixation points carry a lot of information, such as features and the important events happening in the image. They can also reflect on the personality traits of viewers. For this reason, predicting fixation points in an image, i.e. saliency prediction, has been an important research problem for decades.
One of the early works in universal saliency prediction is \cite{itti1998model}
which started an era of several works in this area. With great improvements in the area of deep neural networks, there has been a lot of progress made in this subject (\cite{borji2019saliency}).
While there has been significant progress in the field of saliency prediction, there are still several flaws.
An important concern is the difference of the saliency maps across different individuals or groups. Many factors such as personal features and biases might affect one's fixation points (\cite{pers_on_attention}).
Thus, it is essential to study saliency prediction from a personalized point of view, i.e. personalized saliency prediction.

In this paper, to study saliency prediction from a more personalized perspective, we introduce Clustered Saliency Prediction, which we define as the prediction of saliency maps for groups of similar subjects. We develop methods to first, group subjects based on some personal features and their previous available saliency data (if any exists). Then, considering this clustering of subjects, we learn to predict the saliency maps of subjects.

The first part of our approach for predicting saliency is to find an appropriate clustering of the subjects. We propose the Subject Similarity Clustering (SSC) method, which builds a network using personalized saliency maps of subjects and their personal features. Then, we use the Louvain community detection algorithm on this network to find the clusters. The reason behind using a network structure is to take into account all the factors in relation to each other in a single entity. 
After putting subjects in appropriate clusters, we propose the Multi-Domain Saliency Translation (MDST) model for predicting saliency maps for each cluster. 
We show the improvements of Clustered Saliency Prediction over some existing universal saliency prediction methods in some experiments.
We also propose a method to assign new people to the most appropriate cluster, using their personal features and any of their known saliency maps. We demonstrate this method's effectiveness in choosing the best cluster in our experiments.

There are multiple advantages to Clustered Saliency Prediction over individually personalized saliency prediction. First, by aggregating the saliency maps of similar subjects in clusters, we omit the unimportant noises and catch the main themes in the saliency patterns. 
Also, by putting subjects in clusters based on their saliency maps and personal features, we are actually assigning them to communities and we get better knowledge of the subjects which we can use for further applications, such as recommendation systems. Another advantage is that by putting subjects in clusters, their privacy would be further preserved, since it will be more difficult to associate an output with a specific user. This clustering approach was not used by prior works, and is a contribution of our paper.

In summary, our contributions in this paper are: 1) Proposing a new clustering method, Subject Similarity Clustering, using a network based structure and Louvain community detection algorithm, 2) Proposing the MDST model to convert universal saliency predictions obtained by previous methods, to saliency maps of the clusters, 3) Conducting experiments on a publicly available dataset containing personalized saliency maps, and demonstrating the superiority of our results to some existing saliency prediction methods, 4) Demonstrating the effectiveness of our clustering method in improving prediction of saliency maps, by comparing the results with some baseline cases, 5) Proposing a method to assign new people to their closest cluster, using their personal features and any of their available saliency maps and prove its usefulness.

\section{Related work}

\textbf{Universal saliency prediction.}
Some of the first works in the area of saliency prediction are \cite{itti1998model, treisman1980feature, koch1987shifts}. These early models were mostly based on extracting simple feature maps from the images. In \cite{cerf2009faces}, they have focused on extracting high-level image features such as faces, text elements, etc. and the extent to which they attract attention.
Some of the first Deep Neural Network (DNN) based works on saliency prediction are eDN model (\cite{vig2014large}) and DeepGaze I (\cite{deepgaze1kummerer2015}). In DeepGaze I, they show that deep convolutional networks that have been trained on computer vision tasks such as object detection, boost saliency prediction. In \cite{kummerer2016deepgaze}, authors introduce the DeepGaze II model for universal saliency prediction that uses transfer learning from the VGG-19 network to achieve a good performance. In a more recent work, authors in \cite{deepgazeIIE} proposed DeepGaze IIE as an improvement over DeepGaze II. In DeepGaze IIE, they replaced the VGG19 backbone with ResNet50 features (\cite{resnet}), which provides a big improvement on saliency prediction.
Also, ML-Net (\cite{cornia2016deep}) is a deep-learning based architecture for predicting universal saliency maps. This model uses multi-level features extracted from a Convolutional Neural Network (CNN) and it is end-to-end trainable.
The model SalGAN (\cite{pan2017salgan}), takes advantage of Generative Adversarial Networks (GAN), which consists of a VGG-16 based encoder-decoder model as the generator, and a discriminator.\\
\textbf{Personalized saliency prediction.}
In \cite{beyondijcai2017-543}, authors have produced a dataset of Personalized Saliency Maps (PSMs). We use this dataset in our experiments. In \cite{beyondijcai2017-543}, they model PSMs based on Universal Saliency Maps (USMs) shared by different participants and adopt a multitask CNN framework to estimate the discrepancy between PSMs and USMs.
In \cite{xu2018personalized}, which is an extended version of \cite{beyondijcai2017-543}, they similarly decompose a PSM into a USM predictable by previous saliency detection models and a new discrepancy map across users that characterizes personalized saliency. Then, they present a new solution in addition to their previous work towards predicting such discrepancy maps.
In \cite{lin2018s}, the authors develop Personalized Attention Network (PANet), which contains two streams of CNNs that share common feature extraction layers.
One of the main limitations in personalized saliency prediction is the lack of large saliency datasets for each subject. In \cite{moroto2020few}, the authors proposed few-shot personalized saliency prediction using a small amount of training data based on Adaptive Image Selection (AIS) considering object and visual attention. 
In a more recent work in \cite{ishikawa2021saliency}, the proposed model is composed of universal saliency prediction and personalized gaze probability prediction modules and then the results of these two modules are integrated to generate a final saliency map.

\section{Methods}
In this section, we first describe the dataset and the evaluation metrics that we use in our experiments. After, we describe our clustering method. Then, we talk about the universal saliency prediction methods that we use and our proposed MDST model for obtaining the clusters' saliency maps. To easier understand the approach, we explain its usage on the PSM dataset from \cite{beyondijcai2017-543}, but it can be adapted to other personalized saliency map datasets as well. \\
\textbf{Datasets used in experiments. }In order to test our approaches, we used the dataset collected in \cite{beyondijcai2017-543}, which contains personalized saliency maps. The dataset consists of 1600 images with multiple semantic annotations, which were observed by 30 student participants (14 males, 16 females, aged between 20 and 25).
The dataset also includes a survey to collect each observer’s personal information. Specifically, the following information is collected - the observer’s gender (1D), the preference to objects falling into the fashion category (ring, necklace, etc., 11D), the preference/disgust to colors (red, yellow, etc., 16D), the preference to different sports (football, etc., 11D), and the preference to objects falling into other categories (IT, plant, etc., 4D). 
You can see more details of the dataset in \cite{beyondijcai2017-543}.
Subjects responded to each feature with 0 or 1. This yields a 43-dimensional vector for personal features of each subject in five categories: Gender, Fashion, Color, Sport and Other.
We will refer to this dataset as the \textit{PSM dataset}. \\
\textbf{Evaluation metrics. }For a review of many saliency evaluation metrics, see \cite{bylinskii2018different} and \cite{le2013methods}.
To evaluate the performance of our methods, we use Pearson’s Correlation Coefficient (CC), Similarity (SIM), Normalized Scanpath Saliency (NSS) and Area under ROC Curve (AUC) Judd metrics. 

\subsection{Clustering}
\label{clust_sect}
Here, we focus on creating a network structure based on subjects' information and clustering subjects using this network. To cluster the subjects, we create a weighted complex network of the subjects in the dataset using images seen by the subjects and their personal features. Then, we use a community detection method (\cite{communityPhysRevE.80.056117}) on this weighted network, called the Louvain algorithm (\cite{blondel2008fast}), for dividing the subjects into groups. We call this clustering algorithm \textit{Subject Similarity Clustering (SSC)}. Now, we describe the SSC algorithm.

 First, we initiate a network called \textit{Subject Similarity Network (SSN)} with 30 nodes $(N_1$, $\ldots$, $N_{30})$, corresponding to subjects $(P_1, \ldots, P_{30})$ in the PSM dataset respectively. We denote the weight of the edge between node $u$ and node $v$ as $W(u,v)$, initialized to zero. We denote the personalized saliency map for person $P_i$ and image $x$ as $S_{PSM}(P_i,x)$, which are normalized to have pixel values in [0,1]. All saliency maps are resized to the same size, with resolution $R$.
For personal features of subjects, we denote the feature vector of person $P$ with $F(P)$ and the subset of feature vector for feature category C with $F_{C}(P)$, e.g. $F_{Gender}(P)$. Also, for feature category $C$, $|C|$ denotes the number of features in this category, e.g. $|Gender| = 1$. For feature categories we consider the feature weights $W_{Gender}$, $W_{Fashion}$, $W_{Color}$, $W_{Sport}$ and $W_{Other}$.
Now, for each pair of nodes $N_i$ and $N_j$ in SSN (order of the nodes does not matter, since SSN is undirected), we add an edge between them with the weight:
\begin{equation}
   W(N_i, N_j) =  \frac{m_{i,j}}{\sum_{x\in I_{i,j}} \frac{\lVert S_{PSM}(P_i,x) - S_{PSM}(P_j,x) \rVert _{1} + 1}{R}} + \sum_{C \in FCATS } \frac{W_{C} \times |C|}{ \lVert  F_{C}(P_i) - F_{C}(P_j)\rVert_{1}+1},
\end{equation}
where $m_{i,j}$ is the number of elements of $I_{i,j}$ which is the set of common stimuli images of subjects $N_i$ and $N_j$, also $FCATS = \{Gender$, $Fashion$, $Color$, $Sport$, $Other\}$.
Using the above algorithm for constructing SSN, we get a weighted network in which the weight of the link between two subjects is an indicator of the similarity of their personal features and personalized saliency maps.
For clustering the subjects, we chose the Louvain algorithm, since this algorithm works well with weighted network. The Louvain algorithm is a heuristic algorithm which aims at maximizing modularity in the process of detection of communities. One of the nice features of this algorithm is that it does not require the number of communities or the size of them, before execution.
By applying the Louvain community detection algorithm to SSN, we obtain the clusters. As we will see in \Cref{clust_experim}, we will experiment different values for $W_{Gender}, W_{Fashion}, W_{Color}, W_{Sport}$ and $W_{Other}$ and choose the case which yields the clustering with the \textbf{highest modularity} in the Louvain algorithm.

\subsection{Universal saliency prediction}
\label{univ_section}
For universal saliency prediction, we evaluated various state-of-the-art methods, such as DeepGaze IIE (\cite{deepgazeIIE}), ML-Net (\cite{cornia2016deep}) and SalGAN (\cite{pan2017salgan}). As we will see in \Cref{MDST_section}, we use the USMs obtained by these methods as part of the inputs in training the MDST model.

\subsection{Multi-Domain Saliency Translation}
\label{MDST_section}
In order to obtain the saliency maps for each cluster, we propose the multi-domain saliency translation (MDST) model, based on Conditional Generative Adversarial Networks (cGAN). This model is adapted from the Pix2Pix model by Isola \etal~\cite{isola2017image}. Here, we have a cluster-mapping network inspired by the class network in \cite{hypermoduleLaria} which takes the cluster label as the input and generates a point in the class space. This network is comprised of an embedding layer and followed by 4 fully connected layers. The Equalized Learning Rate technique (\cite{karras2018progressive}) is used for the fully connected layers. The output of the cluster-mapping network for each cluster label is concatenated to the input image and its USM along the channel dimension and fed to the generator which is identical to the U-Net generator of the Pix2Pix model. Also, to the discriminator, which is identical to the discriminator of Pix2Pix model, we feed the concatenation of the input image and its USM and the result generated by the generator, concatenated along the channel dimension. So, this model outputs the Clustered Saliency map for a viewer in cluster $c$ and image $x$, given input image $x$, USM of $x$ using a method of choice and cluster label ($c$) as inputs. The objective of our network, similarly to the one for Pix2Pix model, is as below:
\begin{equation}
G^{\star} = arg \min_G \max_D \mathcal{L}_{cGAN} (G, D) + \lambda \mathcal{L}_{L1(G)},
\label{eq_loss}
\end{equation}
 where G is the generator, D is the discriminator and $\lambda$ is the weight for L1 loss function. We replace the loss function for cGAN, which is a cross-entropy objective in Pix2Pix model, with a Mean Squared Error (MSE) loss. This modification improves the performance based on the results of some experiments. Here we use $\lambda = 100$ in \Cref{eq_loss}. We train the model for 200 epochs, using Adam optimizer with initial learning rate of 0.0002 for the first 100 epochs and linearly decaying learning rate to 0 in the remaining 100 epochs. For the Adam optimizer, the momentum parameters are $\beta_1 = 0.5$ and $\beta_2 = 0.999$ and we use weight decay of 0.00001, to help prevent overfitting.   \\
\textbf{Training approach for MDST model.}
Suppose that after clustering subjects using the SSC algorithm, we have $n$ clusters $C_1, \dots, C_n$. Assume that in the PSM dataset we have personalized saliency maps for a set of stimuli images $I$.
We denote the USM of an image $x$ obtained using method $M$ by $S_{USM}^M(x)$ and define $\{ S_{USM}^M(x') \, | \, x'\in I \}= S_{USM}^M(I)$.
For a cluster $C_i$ and image $x \in I$,
$S_{PSM}(C_i,x) = \frac{\sum_{P \in C_i}S_{PSM}(P,x)}{|C_i|}$, where $|C_i|$ is the number of subjects in $C_i$.
MDST is trained by setting image $x$, $S_{USM}^M(x)$ and label of $C_i$ as input and $S_{PSM}(C_i,x)$ as the target image, for all stimuli images $x \in I$ and all clusters $C_i \in \{C_1, \dots, C_n\}$.

\subsection{Prediction of saliency maps for a new subject}
\label{new_pers_section}

\begin{figure*}
\begin{center}
\includegraphics[scale=0.65]{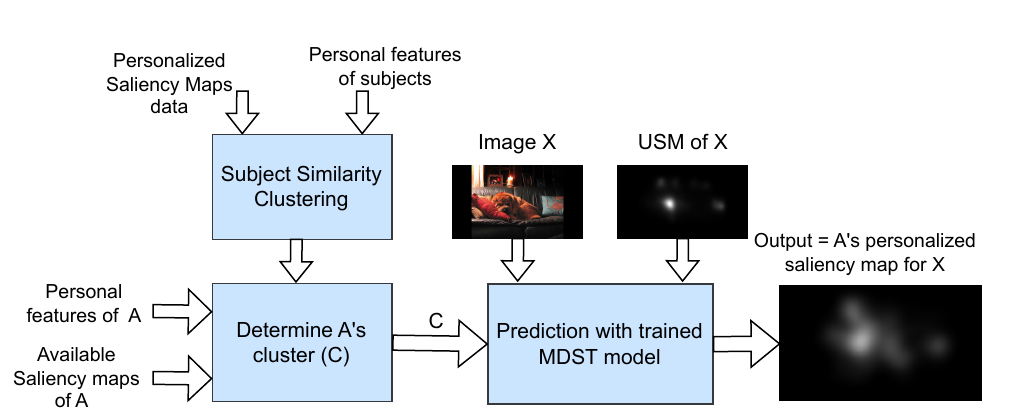}
\end{center}
\caption{ The pipeline of our Clustered Saliency Prediction model, which is combined from clustering of the subjects and the MDST model.}
\label{comb}
\end{figure*}

Having a dataset $D$ that contains personalized saliency data and personal information of some subjects, for a new subject $A$, we want to predict the $A$'s saliency map for an image stimulus. 
First, using the SSC algorithm, we put all the subjects of the dataset $D$ into clusters, $\{C_1, \dots, C_n\}$, and train MDST model, as explained in \Cref{MDST_section}. 
Now we want to determine which cluster in dataset $D$ subject $A$ belongs to, given $A$'s personal features and $A$'s available saliency maps. 
Assume that from $A$, we have the vector $F(A) = [f^A_1,f^A_2,\cdots , f^A_m]$ of values for the set of personal features $\{f_1,f_2,\cdots , f_m \}$, and a set $\{ S_{PSM}(A, x) \, | \, x \in I \} = S_{PSM}(A, I)$ for the set of images, $I$, such that all $x \in I$ exists in the stimuli images set of dataset $D$. 
We normalize all the saliency maps such that their pixel values are in [0,1] and resize them to the same size, with resolution $R$. 
For each subject $P$ in dataset $D$, as the feature vector we only consider the values of the features in $\{f_1,\cdots , f_m \}$, and show this vector by $F(P) = [f^P_1,\cdots , f^P_m]$. For each cluster $C_i$, we denote $F(C_i)$
as the element-wise average of all $F(P)$ for $P \in C_i$.
For feature categories, we consider the same feature weights used for SSC algorithm (e.g. $W_{Gender}$, $W_{Fashion}$, $W_{Color}$, $W_{Sport}$ and $W_{Other}$ for the PSM dataset) as in \Cref{clust_sect}. We denote the subset of features of $F(P)$ and $F(C_i)$
which are in feature category $C$ by $F_C(P)$ and $F_C(C_i)$.
We define a closeness measure between subject $A$ and each cluster $C_i$, called $SalClose(A, C_i)$, computed as below. 
\begin{equation}
    SalClose(A, C_{i}) = \frac{m_{A,C_i}}{\sum_{x\in I_{A,C_i}} \frac{\lVert S_{PSM}(A, x) - S_{PSM}(C_i,x) \rVert _{1} + 1}{R}} + \sum_{C \in FCATS } \frac{W_{C} \times |C|}{ \lVert F_{C}(C_i) - F_{C}(A)\rVert_{1}+1}, 
\end{equation}
where $m_{A,C_i}$ is the number of elements of $I_{A,C_i}$ which is the set of common stimuli images between $A$ and cluster $C_i$ and $FCATS$ is the set of feature categories of dataset $D$, e.g. $\{Gender$, $Fashion$, $Color$, $Sport$, $Other\}$ for PSM dataset.
We assign $A$ to the cluster $C_{ch}$ such that $SalClose(A, C_{ch})$ is the maximum among all the clusters. 
To predict the saliency map of $A$ for an image stimulus $x_s$, we input $x_s$, $S_{USM}^M(x_s)$ ($M$ is a chosen USM prediction model) and label of cluster $C_{ch}$ to MDST model. We consider the output as the Clustered Saliency Prediction for $A$.
The overall pipeline of this process is shown in \Cref{comb}.

\section{Experimental results}
\label{experiment}

\subsection{Clustering}
\label{clust_experim}

For each of the 6 random splits of train/validation/test sets with proportions of 64\%, 16\% and 20\%, for feature weights $W_{Gender}$, $W_{Fashion}$, $W_{Color}$, $W_{Sport}$ and $W_{Other}$, we examine each of the values in \{1, 2, 4, 8\} (1024 total cases) in SSC algorithm in \Cref{clust_sect} using train set images of the split and pick the case which yields the clustering with highest modularity. The average of the obtained feature weights over the 6 splits are $W_{Gender} =3.5$, $W_{Fashion}=8$, $W_{Color}=1$, $W_{Sport}=1$ and $W_{Other}=1$. These settings for each split give 3 clusters which we use in the experiments in \Cref{experiment_mdst}. For the case with $W_{Gender} =0$, $W_{Fashion}=0$, $W_{Color}=0$, $W_{Sport}=0$ and $W_{Other}=0$, we get one cluster with all the subjects in it.

\subsection{Multi-Domain Saliency Translation model}
\label{experiment_mdst}

\begin{figure*}[!t]
\begin{center}

\includegraphics[scale=.55]{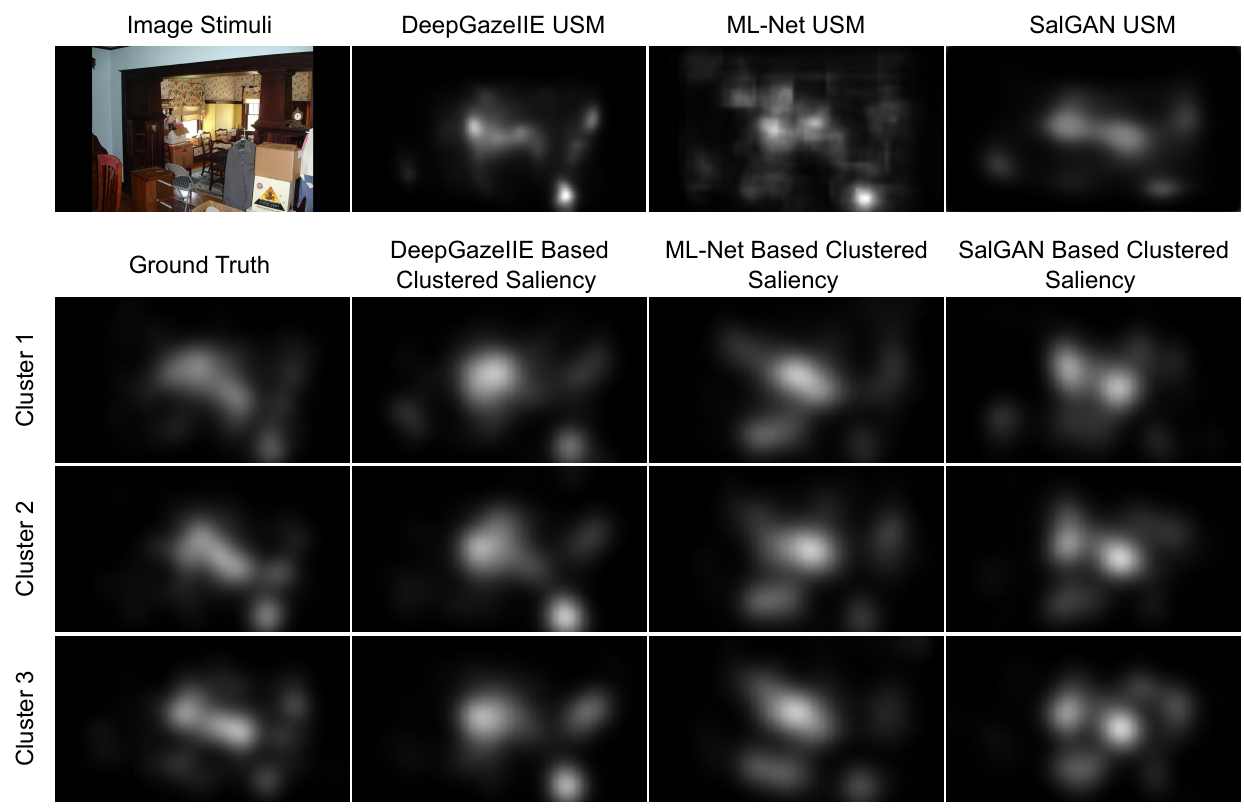}
\end{center}
\caption{Clustered Saliency Prediction for the depicted stimulus image for obtained clusters from \Cref{clust_experim}, using different USM prediction methods as the base for MDST model.}\label{avg_tr_all_fig}
\end{figure*}

We run 3 different set of experiments, where in each set we use one of DeepGaze IIE, ML-Net and SalGAN, to obtain USMs, as a part of the input for MDST model.
We use data augmentation techniques, \emph{i.e.} resize to 286$\times$286 pixels and then random crop to 256$\times$256 pixels and also random horizontal flip for the input images and USMs and PSMs. 
The cluster label is mapped to a code of size 256$\times$16 by an embedding layer and the output is fed into 4 consecutive fully connected layers with the same input and output sizes, 256$\times$16. The output of this is duplicated 4 times, concatenated to each other and resized to shape (256, 256). In all the experiments, we train MDST for 200 epochs (as also explained in \Cref{MDST_section}) and batch size of 16, using the clustering obtained in \Cref{clust_experim}.
The evaluation results are found in \Cref{all_scores_max_modul}, where the results for each metric and each experiment are the average of scores for all subjects in the PSM dataset on the test set over 6 random splits of images into train/validation/test sets with proportions of 64\%, 16\% and 20\% (same splits as in \Cref{clust_experim}). In \Cref{avg_tr_all_fig}, we see the USMs of a random image using each of USM prediction models, and the clustered predictions based on these USMs. 

\begin{table}[]
\begin{center}
    
\begin{tabular}{|c|c|c|c|c|}
\hline
\textbf{Prediction method} & \textbf{CC} & \textbf{SIM} & \textbf{AUC} & \textbf{NSS} \\ \hline
DeepGaze IIE based Clustered      &\textbf{0.7418} &\textbf{0.6369} &0.8862 &2.2518              \\ \hline 
 DeepGaze IIE                      &0.6768 &0.5949 &\textbf{0.8972} &\textbf{2.6413}              \\ \hline
ML-Net based Clustered      &0.7115 &0.6145 &0.8765 &2.1360              \\ \hline 
 ML-Net                      &0.6504 &0.5701 &0.8729 &2.2585              \\ \hline
SalGAN based Clustered      &0.6938 &0.6026 &0.8735 &2.0772              \\ \hline 
 SalGAN                      &0.6606 &0.5816 &0.8757 &2.0182              \\ \hline
\end{tabular}
\end{center}
\caption{\label{all_scores_max_modul}Mean performance of our Clustered Saliency Prediction for all subjects in PSM dataset and comparison to 3 USM prediction methods. Higher score in each metric is better.}
\end{table}

As in \Cref{all_scores_max_modul}, the results for CC and SIM metrics are higher in the Clustered Saliency Predictions, than the results of universal saliency predictions. Also for NSS metric, SalGAN based Clustered Saliency Prediction shows superiority to SalGAN. 
For AUC Judd metric, the performance of ML-Net based Clustered Saliency Prediction is better than ML-Net. All of these show the improvement of our model over universal saliency prediction models. 
The reason for higher performance in AUC Judd and NSS for some cases in the universal saliency model is that the MDST model is trained to convert input images, USMs and cluster labels to saliency maps of the clusters and it does not take into account the unprocessed fixation points data, while AUC Judd and NSS use fixation points as ground truths. Considering this, AUC Judd and NSS might not be good metrics for our evaluations. As we see in \Cref{avg_tr_all_fig}, in the USMs obtained by DeepGaze IIE, the salient parts are more sharp than the salient parts in the Clustered Saliency Predictions for each cluster. Since NSS penalizes false positives at fixation locations, a more sharp saliency map which has higher peaks, might gain a higher NSS than a saliency map with more spread-out salient regions.

To prove the positive effect of the clustering by SSC algorithm on the saliency prediction, we evaluate the results of DeepGaze IIE based MDST network on three baseline cases: 1) having only one cluster comprised of all the subjects, 2) having 3 random clusters, where each person is assigned randomly to one of them, based on uniform distribution, 3) having 30 clusters, where each subject is in a separate cluster. Based on \Cref{ablation_table}, where the same 6 random splits as in \Cref{all_scores_max_modul} were used, for DeepGaze IIE based Clustered Saliency using SSC clustering approach the performance is higher for all the metrics than baseline cases 2 and 3 and is higher in some metrics than baseline case 1. The most populated cluster of each split over 6 random splits, all have the same 10 members in common, plus a few different members for each split. As we see in \Cref{ablation_table}, the average performance for these most populated clusters of all the splits is much improved compared to all other cases. Considering this observation, to get an even further improved performance, for individuals outside of the most-populated cluster we can also use the MDST model trained on the average PSMs of all the individuals (similar to baseline case 1) to predict saliency maps.

    
\begin{table}[]
\begin{center}

\begin{tabular}{|cl|cccc|}
\hline
\multicolumn{1}{|c|}{\textbf{Clustering}}  &                         & \multicolumn{1}{c|}{\textbf{CC}} & \textbf{SIM}    & \textbf{AUC Judd} & \textbf{NSS}    \\ \hline
\multicolumn{1}{|c|}{\multirow{2}{*}{SSC}} & Most populated cluster  & \textbf{0.7573}                  & \textbf{0.6483} & \textbf{0.8933}   & \textbf{2.3376} \\ \cline{2-2}
\multicolumn{1}{|c|}{}                     & Average of all clusters & 0.7418                           & 0.6369          & 0.8862            & 2.2518          \\ \hline
\multicolumn{2}{|c|}{One cluster}                                    & 0.7422                           & 0.6368          & 0.8876            & 2.2519          \\ \hline
\multicolumn{2}{|c|}{3 Random clusters}                              & 0.7416                           & 0.6368          & 0.8864            & 2.2487          \\ \hline
\multicolumn{2}{|c|}{30 clusters}                                    & 0.7274                           & 0.6295          & 0.8687            & 2.2736          \\ \hline
\end{tabular}
\end{center}
\caption{\label{ablation_table} Performance of DeepGaze IIE based MDST network on different ways of clustering, averaged over 6 random splits. "SSC" in the 1st row stands for Subject Similarity Clustering (same results as in the 1st row of \Cref{all_scores_max_modul}). Higher score for each metric is better.}

\end{table}

\subsection{Saliency prediction for new subjects and comparison with other methods}

Similar to \cite{xu2018personalized}, from the PSM dataset we randomly pick 20 subjects whose PSMs are used for clustering and training MDST.
The other 10 subjects are used as new targets.
In open-set evaluation, the model trained on the 20 subjects is tested on PSMs in the test set of the 10 other subjects. In closed-set evaluation, the trained model is tested on PSMs in the test set of the 20 subjects.
Here we have used 5 random splits, where in each split proportions for images in test, validation and train sets are 20\%, 16\% and 64\% respectively. Using the SSC algorithm in \Cref{clust_sect}, from all the combination of weights \{1, 2, 4, 8\} for each feature category, across all 5 splits the chosen feature weights for Gender, Color, Sport, Fashion and Other which yield the clustering with highest modularity are 8, 1, 1, 8, 1, respectively. In each split this gives 2 clusters. Then, we train MDST and evaluate the results (\Cref{comparison_table}). For the 10 test subjects for open-set evaluation, over all 5 splits, we average the scores when assigning each new person to their chosen cluster by our algorithm in \Cref{new_pers_section} and compare to the average of all the scores when assigning each new person to each of the non-chosen clusters. As in \Cref{comparison_table}, the average of scores for assigning the new subjects to chosen clusters for Clustered Saliency Prediction is higher than the average of scores for assigning the new subjects to non-chosen clusters. This proves the effectiveness of our algorithm in assigning the new subjects to the most appropriate cluster for saliency prediction.

In \cite{xu2018personalized}, they predict personalized saliency maps from USMs using multi-task CNN architecture and CNN with Person-specific Information Encoded Filters (CNN-PIEF). They evaluate their approach under closed-set and open-set settings, by randomly choosing 20 subjects to train their models and test on the remaining subjects.
Since we do not know the exact settings of their experiments, comparison with their results under the same circumstances is not possible. However if we overlook this, we have higher performance in CC, AUC Judd and NSS for both DeepGaze IIE based and ML-Net based Clustered Saliency Prediction (\Cref{comparison_table}). 
In \cite{moroto2020few}, they also chose 10 random viewers from the PSM dataset as new targets and used the other 20 subjects for training.
Once again, since the evaluation settings of this paper is not the same as ours, we cannot accurately compare their results to ours. Their performance appears higher than ours (\Cref{comparison_table}), but they are solving a different problem. Keep in mind that our method predicts the average saliency for each cluster, not each person.
\begin{table}[]
\begin{center}
    
\begin{tabular}{|cc|cccc|}
\hline
\multicolumn{2}{|c|}{\textbf{Methods}}                                                                                                                                                                                          & \textbf{CC}     & \textbf{SIM}    & \textbf{AUC Judd} & \textbf{NSS}    \\ \hline
\multicolumn{1}{|c|}{\multirow{2}{*}{\begin{tabular}[c]{@{}c@{}}\cite{xu2018personalized}, \\ closed-set \end{tabular}}} & ML-Net based CNN-PIEF       & 0.6368 & 0.8095 & 0.8365   & 1.5105 \\ \cline{2-6} 
\multicolumn{1}{|c|}{}                                                                                                                        & \begin{tabular}[c]{@{}c@{}}ML-Net based \\ Multi-task CNN\end{tabular} & 0.6463 & 0.8077 & 0.8414   & 1.4960 \\ \hline
\multicolumn{1}{|c|}{\multirow{2}{*}{\begin{tabular}[c]{@{}c@{}}\cite{xu2018personalized}, \\ open-set \end{tabular}}}   & ML-Net based CNN-PIEF      & 0.6450 & 0.8166 & 0.8559   & 1.6879 \\ \cline{2-6} 
\multicolumn{1}{|c|}{}                                                                                                                        & \begin{tabular}[c]{@{}c@{}}ML-Net based \\ Multi-task CNN\end{tabular} & 0.6117 & 0.7946 & 0.8534   & 1.5490 \\ \hline
\multicolumn{1}{|c|}{\cite{moroto2020few}}                                                                                   & Few-shot PSM pred.                                                & 0.7845 & 0.6557 & -        & -      \\ \hline 
\multicolumn{1}{|c|}{\multirow{2}{*}{\begin{tabular}[c]{@{}c@{}}Our method, \\ closed-set \end{tabular}}}                               &ML-Net based Clustered  &0.7107             &0.6167              &0.8725              &2.1057                     \\ \cline{2-6} 
\multicolumn{1}{|c|}{}                                                                                                                        & \begin{tabular}[c]{@{}c@{}}DeepGaze IIE based \\ Clustered \end{tabular}   &0.7417             &0.6398              &0.8819              &2.2181        \\ \hline
\multicolumn{1}{|c|}{\multirow{4}{*}{\begin{tabular}[c]{@{}c@{}}Our method, \\ open-set \end{tabular}}}                                 
& ML-Net based Clustered &0.7030        &0.5981        &0.8852          &2.2019        \\ 
\cline{2-2} 

\multicolumn{1}{|c|}{}
 & \begin{tabular}[c]{@{}c@{}}ML-Net based \\ Non-Chosen Clustered  \end{tabular}  &0.6976        &0.5954        &0.8842          &2.1876        \\ \cline{2-6} 

\multicolumn{1}{|c|}{}                                                                                                                        & \begin{tabular}[c]{@{}c@{}}DeepGaze IIE based \\ Clustered \end{tabular}  &0.7336         &0.6216        &0.8945          &2.3157      \\
\cline{2-2} 
\multicolumn{1}{|c|}{}
 & \begin{tabular}[c]{@{}c@{}}DeepGaze IIE based \\ Non-Chosen Clustered \end{tabular}  &0.7274        &0.6184        &0.8936          &2.3004        \\ \hline
\end{tabular}
\end{center}

\caption{\label{comparison_table} Comparison of our methods under closed-set and open-set evaluation settings with other approaches. The rows "ML-Net/DeepGaze IIE based
Non-Chosen Clustered" have the average for ML-Net/DeepGaze IIE based Clustered Saliency Predictions using MDST, when assigning the new subjects to the clusters \textbf{not chosen} by our algorithm in \Cref{new_pers_section}.}
\end{table}
It is important to keep in mind that our clustered salience approach has other advantages compared to the competing methods mentioned in this section. Most significantly, our method associates individuals with clusters of other viewers. This association can be used to infer additional information for the individual based on any information known for the cluster. For example, it may be known what the favorite music is for members of the cluster, which could be applied to provide recommendations to the new individual. We do not investigate leveraging the cluster associations in this paper, but leave it for future work.

\section{Discussion and Conclusion} 

From our experiments, we conclude that our Clustered Saliency Prediction method improves on standard universal saliency prediction methods. Based on \Cref{all_scores_max_modul}, the performance of DeepGaze IIE based Clustered Saliency Prediction is higher than ML-Net and SalGAN based Clustered Saliency Predictions. This is logical, since DeepGaze IIE has a better performance comparing to two other universal saliency models.  
One of the advantages of our Clustered Saliency Prediction model is that the USMs that we use as the base can be obtained with any universal saliency model. So as universal saliency models improve, we can upgrade our model by using a better performing universal saliency model base.

 Also, as we discussed in \Cref{experiment_mdst}, our SSC approach divides the subjects into clusters, where each cluster contains subjects with more similar saliency patterns. We also see that some features categories such as fashion, appear to correlate more with the saliency of subjects. Moreover, the clusters found in our approach, can be used in future works for further applications, such as recommendation systems, advertising, etc. 

\section{Acknowledgement}
We acknowledge the support of the Natural Sciences and Engineering Research Council of Canada (NSERC) and funding from the Ministère de l'Économie, de l'Innovation et de l'Énergie of Québec.

\bibliography{egbib}

\begin{thebibliography}{25}
\providecommand{\natexlab}[1]{#1}
\providecommand{\url}[1]{\texttt{#1}}
\expandafter\ifx\csname urlstyle\endcsname\relax
  \providecommand{\doi}[1]{doi: #1}\else
  \providecommand{\doi}{doi: \begingroup \urlstyle{rm}\Url}\fi

\bibitem[Blondel et~al.(2008)Blondel, Guillaume, Lambiotte, and Lefebvre]{blondel2008fast}
Vincent~D Blondel, Jean-Loup Guillaume, Renaud Lambiotte, and Etienne Lefebvre.
\newblock Fast unfolding of communities in large networks.
\newblock \emph{Journal of statistical mechanics: theory and experiment}, 2008\penalty0 (10):\penalty0 P10008, 2008.

\bibitem[Borji(2019)]{borji2019saliency}
Ali Borji.
\newblock Saliency prediction in the deep learning era: Successes and limitations.
\newblock \emph{IEEE transactions on pattern analysis and machine intelligence}, 2019.

\bibitem[Bylinskii et~al.(2018)Bylinskii, Judd, Oliva, Torralba, and Durand]{bylinskii2018different}
Zoya Bylinskii, Tilke Judd, Aude Oliva, Antonio Torralba, and Fr{\'e}do Durand.
\newblock What do different evaluation metrics tell us about saliency models?
\newblock \emph{IEEE transactions on pattern analysis and machine intelligence}, 41\penalty0 (3):\penalty0 740--757, 2018.

\bibitem[Cerf et~al.(2009)Cerf, Frady, and Koch]{cerf2009faces}
Moran Cerf, E~Paxon Frady, and Christof Koch.
\newblock Faces and text attract gaze independent of the task: Experimental data and computer model.
\newblock \emph{Journal of vision}, 9\penalty0 (12):\penalty0 10--10, 2009.

\bibitem[Cornia et~al.(2016)Cornia, Baraldi, Serra, and Cucchiara]{cornia2016deep}
Marcella Cornia, Lorenzo Baraldi, Giuseppe Serra, and Rita Cucchiara.
\newblock A deep multi-level network for saliency prediction.
\newblock In \emph{2016 23rd International Conference on Pattern Recognition (ICPR)}, pages 3488--3493. IEEE, 2016.

\bibitem[He et~al.(2016)He, Zhang, Ren, and Sun]{resnet}
Kaiming He, Xiangyu Zhang, Shaoqing Ren, and Jian Sun.
\newblock Deep residual learning for image recognition.
\newblock In \emph{2016 IEEE Conference on Computer Vision and Pattern Recognition (CVPR)}, pages 770--778, 2016.
\newblock \doi{10.1109/CVPR.2016.90}.

\bibitem[Ishikawa and Yakoh(2021)]{ishikawa2021saliency}
Tomoki Ishikawa and Takahiro Yakoh.
\newblock Saliency prediction based on object recognition and gaze analysis.
\newblock \emph{Electronics and Communications in Japan}, 104\penalty0 (2):\penalty0 e12303, 2021.

\bibitem[Isola et~al.(2017)Isola, Zhu, Zhou, and Efros]{isola2017image}
Phillip Isola, Jun-Yan Zhu, Tinghui Zhou, and Alexei~A Efros.
\newblock Image-to-image translation with conditional adversarial networks.
\newblock In \emph{Proceedings of the IEEE conference on computer vision and pattern recognition}, pages 1125--1134, 2017.

\bibitem[Itti et~al.(1998)Itti, Koch, and Niebur]{itti1998model}
Laurent Itti, Christof Koch, and Ernst Niebur.
\newblock A model of saliency-based visual attention for rapid scene analysis.
\newblock \emph{IEEE Transactions on pattern analysis and machine intelligence}, 20\penalty0 (11):\penalty0 1254--1259, 1998.

\bibitem[Karras et~al.(2018)Karras, Aila, Laine, and Lehtinen]{karras2018progressive}
Tero Karras, Timo Aila, Samuli Laine, and Jaakko Lehtinen.
\newblock Progressive growing of {GAN}s for improved quality, stability, and variation.
\newblock In \emph{International Conference on Learning Representations}, 2018.
\newblock URL \url{https://openreview.net/forum?id=Hk99zCeAb}.

\bibitem[Koch and Ullman(1987)]{koch1987shifts}
Christof Koch and Shimon Ullman.
\newblock Shifts in selective visual attention: towards the underlying neural circuitry.
\newblock In \emph{Matters of intelligence}, pages 115--141. Springer, 1987.

\bibitem[K{\"u}mmerer et~al.(2014)K{\"u}mmerer, Theis, and Bethge]{deepgaze1kummerer2015}
Matthias K{\"u}mmerer, Lucas Theis, and Matthias Bethge.
\newblock Deep gaze {I:} boosting saliency prediction with feature maps trained on imagenet.
\newblock \emph{CoRR}, abs/1411.1045, 2014.

\bibitem[K{\"u}mmerer et~al.(2016)K{\"u}mmerer, Wallis, and Bethge]{kummerer2016deepgaze}
Matthias K{\"u}mmerer, Thomas S.~A. Wallis, and Matthias Bethge.
\newblock Deepgaze {II:} reading fixations from deep features trained on object recognition.
\newblock \emph{ArXiv}, abs/1610.01563, 2016.

\bibitem[Lancichinetti and Fortunato(2009)]{communityPhysRevE.80.056117}
Andrea Lancichinetti and Santo Fortunato.
\newblock Community detection algorithms: A comparative analysis.
\newblock \emph{Phys. Rev. E}, 80:\penalty0 056117, Nov 2009.
\newblock \doi{10.1103/PhysRevE.80.056117}.
\newblock URL \url{https://link.aps.org/doi/10.1103/PhysRevE.80.056117}.

\bibitem[Laria et~al.(2022)Laria, Wang, van~de Weijer, and Raducanu]{hypermoduleLaria}
Héctor Laria, Yaxing Wang, Joost van~de Weijer, and Bogdan Raducanu.
\newblock Transferring unconditional to conditional gans with hyper-modulation.
\newblock In \emph{2022 IEEE/CVF Conference on Computer Vision and Pattern Recognition Workshops (CVPRW)}, pages 3839--3848, 2022.
\newblock \doi{10.1109/CVPRW56347.2022.00429}.

\bibitem[Le~Meur and Baccino(2013)]{le2013methods}
Olivier Le~Meur and Thierry Baccino.
\newblock Methods for comparing scanpaths and saliency maps: strengths and weaknesses.
\newblock \emph{Behavior research methods}, 45\penalty0 (1):\penalty0 251--266, 2013.

\bibitem[Lin and Hui(2018)]{lin2018s}
Sikun Lin and Pan Hui.
\newblock Where's your focus: Personalized attention.
\newblock \emph{arXiv preprint arXiv:1802.07931}, 2018.

\bibitem[Linardos et~al.(2021)Linardos, K{\"{u}}mmerer, Press, and Bethge]{deepgazeIIE}
Akis Linardos, Matthias K{\"{u}}mmerer, Ori Press, and Matthias Bethge.
\newblock Deepgaze {IIE:} calibrated prediction in and out-of-domain for state-of-the-art saliency modeling.
\newblock In \emph{2021 {IEEE/CVF} International Conference on Computer Vision, {ICCV} 2021, Montreal, QC, Canada, October 10-17, 2021}, pages 12899--12908. {IEEE}, 2021.
\newblock \doi{10.1109/ICCV48922.2021.01268}.
\newblock URL \url{https://doi.org/10.1109/ICCV48922.2021.01268}.

\bibitem[Moroto et~al.(2020)Moroto, Maeda, Ogawa, and Haseyama]{moroto2020few}
Yuya Moroto, Keisuke Maeda, Takahiro Ogawa, and Miki Haseyama.
\newblock Few-shot personalized saliency prediction based on adaptive image selection considering object and visual attention.
\newblock \emph{Sensors}, 20\penalty0 (8):\penalty0 2170, 2020.

\bibitem[Pan et~al.(2017)Pan, Canton{-}Ferrer, McGuinness, O'Connor, Torres, Sayrol, and Gir{\'{o}}{-}i{-}Nieto]{pan2017salgan}
Junting Pan, Cristian Canton{-}Ferrer, Kevin McGuinness, Noel~E. O'Connor, Jordi Torres, Elisa Sayrol, and Xavier Gir{\'{o}}{-}i{-}Nieto.
\newblock Salgan: Visual saliency prediction with generative adversarial networks.
\newblock \emph{CoRR}, abs/1701.01081, 2017.
\newblock URL \url{http://arxiv.org/abs/1701.01081}.

\bibitem[Treisman and Gelade(1980)]{treisman1980feature}
Anne~M Treisman and Garry Gelade.
\newblock A feature-integration theory of attention.
\newblock \emph{Cognitive psychology}, 12\penalty0 (1):\penalty0 97--136, 1980.

\bibitem[Vig et~al.(2014)Vig, Dorr, and Cox]{vig2014large}
Eleonora Vig, Michael Dorr, and David Cox.
\newblock Large-scale optimization of hierarchical features for saliency prediction in natural images.
\newblock In \emph{Proceedings of the IEEE conference on computer vision and pattern recognition}, pages 2798--2805, 2014.

\bibitem[Wu et~al.(2014)Wu, Bischof, Anderson, Jakobsen, and Kingstone]{pers_on_attention}
David W.-L. Wu, Walter~F. Bischof, Nicola~C. Anderson, Tanya Jakobsen, and Alan Kingstone.
\newblock The influence of personality on social attention.
\newblock \emph{Personality and Individual Differences}, 60:\penalty0 25--29, 2014.
\newblock ISSN 0191-8869.
\newblock \doi{https://doi.org/10.1016/j.paid.2013.11.017}.
\newblock URL \url{https://www.sciencedirect.com/science/article/pii/S0191886913013755}.

\bibitem[Xu et~al.(2017)Xu, Li, Wu, Yu, and Gao]{beyondijcai2017-543}
Yanyu Xu, Nianyi Li, Junru Wu, Jingyi Yu, and Shenghua Gao.
\newblock Beyond universal saliency: Personalized saliency prediction with multi-task cnn.
\newblock In \emph{Proceedings of the Twenty-Sixth International Joint Conference on Artificial Intelligence, {IJCAI-17}}, pages 3887--3893, 2017.
\newblock \doi{10.24963/ijcai.2017/543}.
\newblock URL \url{https://doi.org/10.24963/ijcai.2017/543}.

\bibitem[Xu et~al.(2018)Xu, Gao, Wu, Li, and Yu]{xu2018personalized}
Yanyu Xu, Shenghua Gao, Junru Wu, Nianyi Li, and Jingyi Yu.
\newblock Personalized saliency and its prediction.
\newblock \emph{IEEE transactions on pattern analysis and machine intelligence}, 41\penalty0 (12):\penalty0 2975--2989, 2018.

\end{thebibliography}
\end{document}